\documentclass[lettersize,journal]{IEEEtran}
\usepackage{amsmath,amsfonts}
\usepackage{algorithmic}
\usepackage{algorithm}
\usepackage{array}
\usepackage[caption=false,font=normalsize,labelfont=sf,textfont=sf]{subfig}
\usepackage{textcomp}
\usepackage{stfloats}
\usepackage{url}
\usepackage{verbatim}
\usepackage{graphicx}
\usepackage{cite}
\usepackage{multirow, xcolor, colortbl}  
\usepackage{amsmath,amsfonts,amssymb}
\newcommand{\indicator}[1]{\mathbf{1}_{\{#1\}}}
\usepackage[table]{xcolor} 
\usepackage{pifont}
\usepackage{booktabs}
\newcommand{\cmark}{\ding{51}} 
\newcommand{\xmark}{\ding{55}} 
\begin{document}
\title{DynProto: Dynamic Prototype Evolution for OOD Detection}

\author{Yanqi Wu, Xinhua Lu, Runhe Lai, Qichao Chen, Jia-Xin Zhuang, Wei-Shi Zheng, and Ruixuan Wang
\thanks{Yanqi Wu, Xinhua Lu, Runhe Lai, Wei-Shi Zheng, and Ruixuan Wang are with the School of Computer Science and Engineering, Sun Yat-sen University, Guangzhou 510275, China, also with Peng Cheng Laboratory, Shenzhen 518066, China, and also with the Key Laboratory of Machine Intelligence and Advanced Computing, MOE, Guangzhou 510275, China. (e-mail: wuyq268@mail2.sysu.edu.cn; luxh55@mail2.sysu.edu.cn; lairh5@mail2.sysu.edu.cn; wszheng@ieee.org;  wangruix5@ mail.sysu.edu.cn)}
\thanks{Qichao Chen is with the University of Nottingham, Malaysia.(e-mail: hcxqc1@nottingham.my)}
\thanks{Jia-Xin Zhuang is with the Hong Kong University of Science and Technology, Hong kong, China.(e-mail: jzhuangad@cse.ust.hk)}
\thanks{Corresponding authors: Jia-Xin Zhuang and Ruixuan Wang.}}

\markboth{IEEE TRANSACTIONS ON MULTIMEDIA}%
{Shell \MakeLowercase{\textit{et al.}}: A Sample Article Using IEEEtran.cls for IEEE Journals}


\maketitle

\begin{abstract}
Recent vision-language model (VLM)-based OOD detection methods leverage large-scale corpora to construct negative labels as predefined OOD prototypes, achieving strong detection performance. However, such static textual prototypes are inherently constrained by the coverage of the selected label space and may fail to represent previously unseen or distribution-specific OOD patterns encountered in real-world environments. To overcome this limitation, we propose DynProto, a training-free and architecture-agnostic framework that dynamically models the evolving OOD distribution at test time. Motivated by the observation that OOD samples associated with the same predicted in-distribution (ID) class tend to form compact groups in the feature space, DynProto uses easily detected OOD candidates as anchors to identify harder yet visually related OOD samples. Specifically, the Coarse OOD Pattern Capturing Module maintains class-aware caches to collect candidate OOD features during inference, while the Fine-grained OOD Pattern Refinement Module clusters the cached features and aggregates each cluster into a representative OOD prototype. Together with ID prototypes constructed from the training data, these dynamically evolved OOD prototypes provide explicit references for prototype-based OOD scoring. DynProto requires neither external OOD labels nor additional model training and can be seamlessly applied to both unimodal and vision-language backbones. Extensive experiments across multiple OOD benchmarks demonstrate that DynProto consistently outperforms state-of-the-art methods. On the ImageNet OOD benchmark, DynProto reduces FPR95 by 11.60\% and improves AUROC by 4.70\%.
\end{abstract}

\section{Introduction}
\label{sec:intro}

\IEEEPARstart{D}{eep} neural networks (DNNs) have achieved outstanding success across a wide range of challenging tasks. However, their reliability deteriorates notably in open-world scenarios, where they are exposed to samples from previously unseen classes. In such cases, DNNs often produce overconfident predictions on OOD inputs, which can lead to serious consequences 
in safety-critical domains such as autonomous driving and medical diagnosis.

\begin{figure}[!t]
    \centering
    \includegraphics[width=0.5\textwidth]{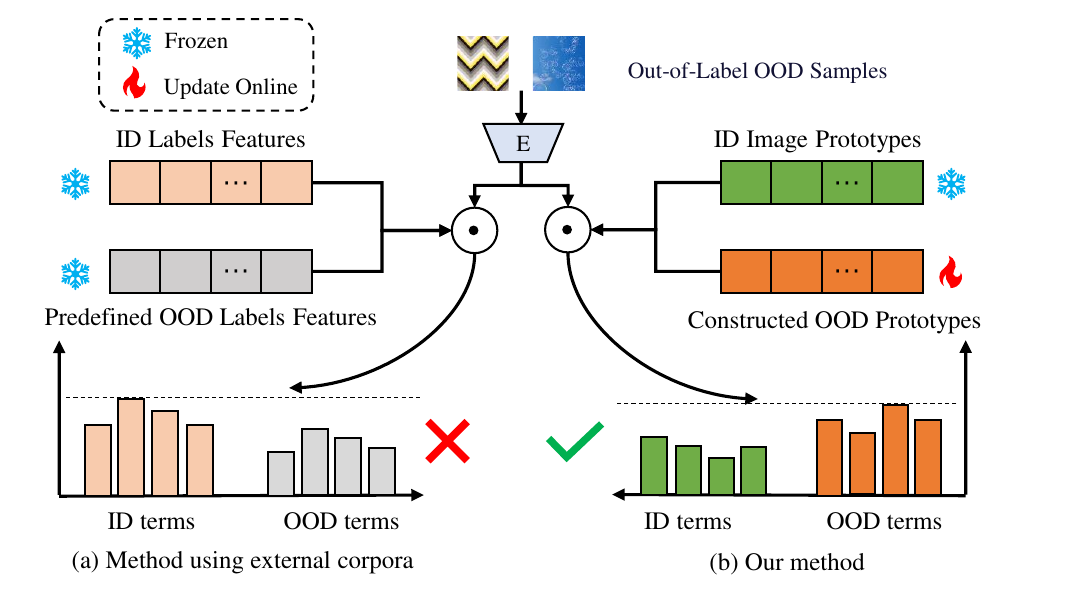}
    \vspace{-2em}
    \caption{Comparison of (a) methods using external corpora, with (b) our proposed DynProto. In the presence of OOD samples beyond the predefined label space, method (a) becomes ineffective, while DynProto dynamically captures emerging OOD patterns.}
    \label{fig1:high_level}
    \vspace{-1.5em}
\end{figure}

Recent studies~\cite{zoc,neglabel,csp} have explored leveraging large-scale corpora or large language models to generate candidate OOD labels. These methods treat the text embeddings of constructed labels as potential OOD prototypes and predict test samples that are similar to these prototypes as OOD class, achieving promising detection performance. However, since their effectiveness heavily depends on the quality of the constructed OOD prototypes, when OOD samples encountered during testing go beyond the coverage of these predefined labels, the effectiveness of such methods deteriorates significantly. Moreover, some of these methods carefully design potential OOD labels for a specific ID dataset and once the ID dataset changes, substantial effort is required to redesign the labels.

To address these issues, we propose DynProto, a dynamic OOD prototype learning framework that captures real OOD feature distributions using only ID information and unlabeled test-time samples, without relying on predefined OOD labels. In our preliminary work, DCAC~\cite{dcac}, we observed that OOD samples associated with the same predicted ID class often share highly similar visual features. DCAC exploits this class-conditioned correlation by storing uncertain test samples in class-aware caches and using them to calibrate overconfident model predictions. However, the cached samples are used only as individual calibration signals, without explicitly modeling the diverse OOD patterns emerging during testing. Building upon this observation, we further find that these class-conditioned OOD samples tend to form compact clusters in the feature space. Although some samples in these clusters can be readily detected as OOD, their visually similar counterparts may still receive high ID confidence. This motivates us to organize easily detected OOD candidates into representative prototypes and use them as anchors to identify harder-to-detect OOD samples with similar visual characteristics.


Building upon this insight, we aim to transform the class-aware test samples collected during inference into explicit representations of the evolving OOD distribution. Specifically, readily detected OOD candidates are used as anchors to identify visually similar yet harder-to-detect OOD samples. To achieve this goal, DynProto comprises two modules. The \textit{Coarse OOD Pattern Capturing Module} maintains a separate cache for each ID class and dynamically collects potential OOD features that are associated with that class but receive low confidence. The \textit{Fine-grained OOD Pattern Refinement Module} then discovers the underlying structure of each cache by clustering the collected features and aggregating each cluster into a representative OOD prototype. This coarse-to-fine procedure reduces redundancy and noise in the cached samples while preserving diverse OOD patterns.
Meanwhile, we construct class-wise ID prototypes from the ID training data. For each test sample, DynProto determines its distributional membership by measuring its relative similarity to the ID prototypes and the dynamically evolved OOD prototypes. As shown in Fig~\ref{fig1:high_level}, unlike methods that rely on static OOD prototypes derived from external textual labels, DynProto directly models the actual OOD patterns emerging in the test stream and continuously adapts its prototypes to the evolving test distribution.

DynProto offers several practical advantages over existing methods based on predefined textual OOD labels. First, it requires neither external OOD labels nor auxiliary outlier data, and introduces no additional model training. Instead, it dynamically learns visual OOD prototypes directly from unlabeled test-time samples. Second, because DynProto operates entirely in the visual feature space, it is not restricted to VLMs equipped with text encoders and can be readily applied to both VLMs and unimodal vision models. Third, DynProto is a flexible plug-and-play framework that can be built upon various existing OOD detectors. These detectors serve as base detectors during the early stage of testing to collect reliable OOD candidates, while the dynamically constructed prototypes progressively provide more informative references as testing proceeds. Consequently, DynProto can benefit from stronger base detectors while remaining compatible with a broad range of model architectures and OOD scoring functions. Our main contributions are summarized as follows:
\begin{itemize}
\item Building upon our previous observation that OOD samples associated with the same predicted ID class exhibit strong visual correlations, we further show that they form compact patterns in the feature space, enabling easily detected OOD samples to help identify similar hard cases.

\item We propose DynProto, a training-free and architecture-agnostic framework that dynamically constructs visual OOD prototypes through coarse-to-fine test-time pattern collection and refinement, without relying on external OOD labels.

\item Extensive experiments across multiple benchmarks demonstrate the effectiveness of DynProto. On the ImageNet OOD benchmark with CLIP-B/16, it reduces FPR95 by 11.60\% and improves AUROC by 4.70\% over the strongest baseline.

\end{itemize}
\section{Related Work}
\label{sec:related_work}

\noindent\textbf{Prototype Learning.}
A prototype is a representative feature embedding that characterizes a class or a local data distribution and is commonly used for similarity-based classification and representation learning. Recently, prototype learning has been introduced into OOD detection. Hu et al.~\cite{richsemantics} introduce knowledge-guided virtual classes and transfer external semantic structures into the visual prototype space, enabling more discriminative prototype learning for open-set action recognition.
SIREN~\cite{siren} shapes hyperspherical feature representations using von Mises--Fisher prototypes to improve the separation between ID and OOD samples. CIDER~\cite{cider} encourages compact intra-class and dispersed inter-class representations around class prototypes. PALM~\cite{PALM2024} models each class with multiple prototypes to capture intra-class diversity. DHE~\cite{DHE} optimizes class prototypes on a hypersphere to obtain provable discrimination between ID and OOD samples, while SPROD~\cite{SPROD} refines ID prototypes to mitigate the influence of spurious correlations. These methods mainly focus on learning more representative ID prototypes. Although POP~\cite{POP} introduces virtual OOD prototypes to reshape the decision boundary, such synthetic prototypes do not explicitly represent the OOD patterns encountered in real test environments. In contrast, DynProto dynamically constructs visual OOD prototypes from unlabeled test-time samples, enabling the model to capture emerging OOD patterns without additional training.

\noindent\textbf{OOD Detection Based on External Textual Information.} 
Benefiting from CLIP’s~\cite{clip} strong generalization capability in open-world scenarios, many recent methods have leveraged its rich textual knowledge by obtaining potential OOD text labels from external text corpora or large language models to assist CLIP in OOD detection. ZOC~\cite{zoc} trains a captioner on external datasets to generate potential OOD text labels. NegLabel~\cite{neglabel} incorporates the WordNet~\cite{wordnet} to mine semantically unrelated words to the ID labels. CSP~\cite{csp} enhances the activation probability of labels by constructing a set of low-correlation super-class labels. NegRefine~\cite{negrefine} refines negative-label-based zero-shot OOD detection by filtering semantically ambiguous labels and introducing a multi-matching-aware scoring strategy. SynOOD~\cite{synood} employs a diffusion-based in-painting pipeline to generate near-boundary OOD images, and then finetunes the CLIP model by aligning those synthetic samples with negative text-label features drawn from an external label space beyond the ID classes.

\noindent\textbf{Test-Time OOD Detection.}
Recent studies have explored using unlabeled test-time samples to improve OOD detection under evolving environments. AUTO~\cite{auto} adapts the model parameters through stochastic gradient descent to reduce overconfidence on potential OOD samples. Yu et al.~\cite{selflabeling} propose a self-labeling framework that combines prototypical contrastive learning and mutual information maximization for few-shot open-set domain adaptation. RTL++~\cite{rtl} models a linear relationship between OOD scores and visual features, but this assumption may become unreliable when the feature distributions of ID and near-OOD samples overlap. AdaND~\cite{adand} freezes the backbone and trains a lightweight noise detector to identify potential OOD samples. TULIP~\cite{tulip} incorporates prior knowledge of ID and OOD distributions to calibrate predictive uncertainty at test time. OODD~\cite{oodd} maintains a priority queue of representative OOD features and uses them to calibrate the OOD scores of incoming samples. TTL~\cite{TTL} uses pseudo-labeled test samples to optimize class-specific learnable OOD prompts, purifies noisy OOD knowledge, and maintains a textual knowledge bank for cross-batch score calibration. AdaNeg~\cite{adaneg} also constructs test-time memory banks, but each memory bank is associated with a predefined negative textual label. Therefore, its performance remains dependent on external OOD labels and its applicability is limited to VLMs. In contrast, DynProto constructs class-aware visual memories without predefined negative labels and dynamically refines them into OOD prototypes, making it compatible with both unimodal and multimodal architectures. 
In our preliminary work, DCAC~\cite{dcac} maintains class-aware caches of uncertain test samples and uses their visual features and predictive distributions to calibrate overconfident model outputs. While DCAC demonstrates the effectiveness of exploiting class-conditioned correlations among test samples, the cached samples primarily serve as instance-level calibration signals and are not organized into explicit representations of the underlying OOD distribution. DynProto extends this idea by clustering class-aware cached features into representative OOD prototypes and performing detection through the relative similarity between test samples and ID/OOD prototypes.

\section{Methodology}

\begin{figure*}[t]
\centering
\includegraphics[width=\textwidth]{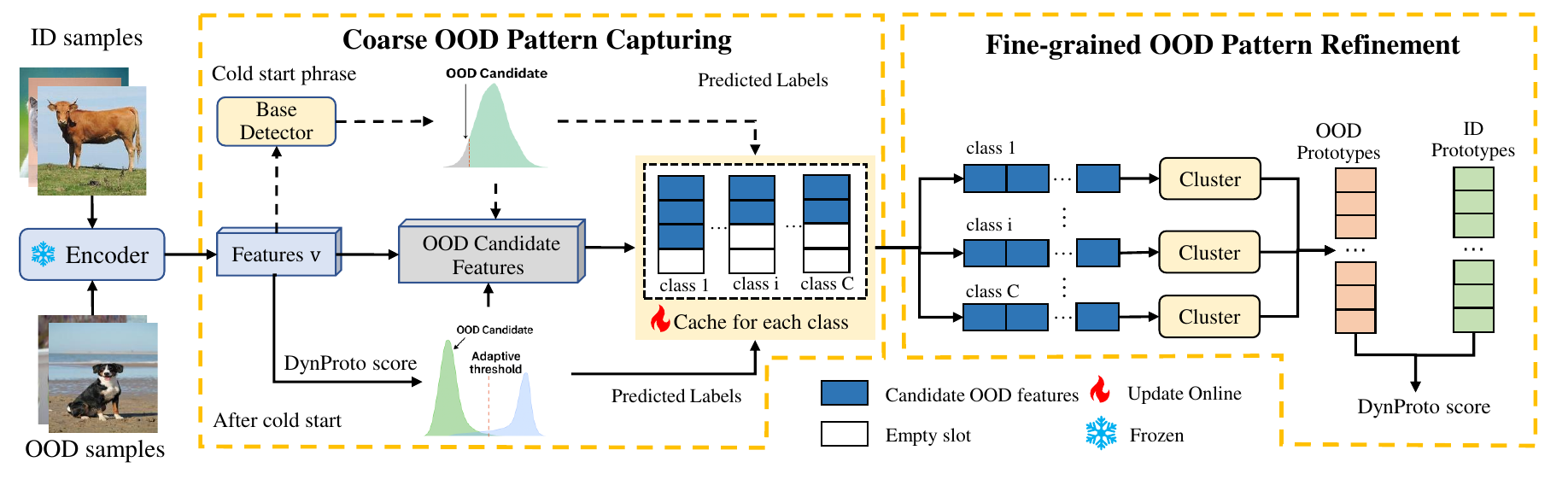} 
\vspace{-2em}
\caption{Overview of DynProto. The Coarse OOD Pattern Capturing Module stores candidate OOD features that are predicted to each ID class. In the Fine-grained OOD Pattern Refinement Module, cached features are clustered into representative OOD prototypes. Meanwhile, ID prototypes are constructed from the training set. During testing, the OOD score for each sample is calculated using the DynProto score from Eq.~\ref{s_proto}, reflecting the similarity of the sample’s feature to both ID and dynamically learned OOD prototypes.}
\label{fig:framework}
\vspace{-1.5em}
\end{figure*}

\subsection{Preliminaries}
\noindent\textbf{OOD Detection.} 
Generally, OOD detection can be regarded as a binary classification problem, in which the model aims to determine whether an input image $\mathbf{x}$ belongs to the ID or OOD.
Formally, the OOD detection task can be defined as:
\begingroup
\small
\begin{equation}
D(\mathbf{x}) = 
\begin{cases} 
1, & \text{if } S(\mathbf{x}) \geq \gamma \\
0, & \text{otherwise}
\end{cases}\;\;\;,
\label{eq:ood-detection}
\end{equation}
\endgroup
where $D(\cdot)$ denotes the model (\textit{i.e.}, an OOD detector) trained on the ID training dataset, $\mathbf{x}$ is sampled from a mixture of the ID and OOD test datasets, 1 and 0 denote ID and OOD classes, respectively, $S(\cdot)$ represents the OOD detection score function and $\gamma$ serves as a threshold hyperparameter. ID samples are expected to receive higher scores than OOD samples.

\noindent\textbf{NegLabel}~\cite{neglabel} extends the CLIP framework to OOD detection by introducing negative labels that serve as semantic prototypes for OOD data. Given the original ID label set $\mathcal{Y}=\{y_1,...,y_C\}$, NegLabel augments it with a set of negative labels $\mathcal{Y}^-=\{y_{C+1},...y_{C+M}\}$ sampled from broad textual corpora, ensuring $\mathcal{Y}\cap\mathcal{Y}^-=\varnothing$. Using the CLIP text encoder, the full text feature matrix is obtained as: 
\begingroup
\small
\begin{equation}
P = f_{\text{txt}}\bigl(\rho(\mathcal{Y} \cup \mathcal{Y}^-)\bigr) \in \mathbb{R}^{(R+M)\times D},
\end{equation}
\endgroup
where $\rho(\cdot)$ denotes the text prompt (e.g., \textit{``the nice $<$label$>$''}). Given an image feature $\mathbf{v} = f_{\mathrm{img}}(\mathrm{x}) \in \mathbb{R}^D$, the OOD score is defined as the cumulative probability of ID classes:
\begingroup
\small
\begin{equation}
\label{neglabel_score}
S_{\mathrm{nl}}(\mathbf{v}) = 
\frac{
    \sum_{i=1}^{C} e^{\cos(\mathbf{v}, \mathbf{p}_i) / \tau}
}{
    \sum_{i=1}^{C} e^{\cos(\mathbf{v}, \mathbf{p}_i) / \tau}
    + \sum_{j=C+1}^{C+M} e^{\cos(\mathbf{v}, \mathbf{p}_j) / \tau}
},
\end{equation}
\endgroup
where $\tau$ is the temperature parameter. $\mathbf{p}_i$ and $\mathbf{p}_j$ respectively represent the ID text feature and the negative label feature in $P$. A lower $S_{\mathrm{nl}}(\mathbf{v})$ indicates that the image is more semantically aligned with the negative labels, suggesting a higher likelihood of being OOD sample. In our method, we replace $\mathbf{p}_i$ and $\mathbf{p}_j$ with the ID image prototypes and our constructed OOD image prototypes, respectively.

\subsection{Overview}
As shown in Figure~\ref{fig:framework}, we propose DynProto, a dynamic prototype learning framework for OOD detection.
Unlike existing methods that rely on pre-defined OOD labels, DynProto constructs class-aware visual OOD prototypes dynamically during testing, only using ID-related information.
We design two modules. The \textit{Coarse OOD Pattern Capturing Module} (COPC, Section~\ref{sec:coarse}) maintains a class-wise cache to collect test samples predicted as each ID class with low OOD scores, and the \textit{Fine-grained OOD Pattern Refinement Module} (FOPR, Section~\ref{sec:fine}) performs clustering on the cached samples and aggregates the features within each cluster to form an OOD prototype, enabling fine-grained representation of OOD patterns.
After that, we use DynProto score (Eq.~\ref{s_proto}) to determine whether a sample belongs to the ID. Finally, we discuss the cold start strategy of DynProto (Section~\ref{sec:cold_start}).
During the cold start phase, when the caches contain only a few samples, DynProto relies on a base OOD detector for sample collection.
As testing proceeds and more diverse samples are collected, we use DynProto score with an adaptive collecting threshold for sample collection.

\subsection{Coarse OOD Pattern Capturing}
\label{sec:coarse}
To construct OOD prototypes, we first need to collect diverse OOD patterns. We maintain a cache for each ID class, which is initially empty. Given an input $\textbf{x}$, we feed it into detector and get an OOD score. If the score of $\textbf{x}$ is lower, it is more likely to be an OOD sample. Therefore, if the score of $\textbf{x}$ is lower than a predefined threshold, $\textbf{x}$ is regarded as a potential OOD sample and its feature is stored in the cache associated with the  predicted ID class.

To maintain efficiency, each cache is assigned a fixed capacity and updated using a first-in-first-out (FIFO) strategy. When the cache is not full, the visual feature $\mathbf{v}$ of an OOD candidate is directly stored in the cache of its predicted class. If the cache is full, the oldest entry is removed before adding the new one. Based on temporal locality, recent samples are more likely to be correlated with the current test sample. Therefore, the FIFO strategy helps retain samples most relevant to the current testing environment while discarding outdated ones, enhancing cache efficiency. More details about the impact of cache update strategy are provided in Section~\ref{Update_Strategies}.

\subsection{Fine-grained OOD Pattern Refinement}
\label{sec:fine}
After cache updating, we use the collected samples to construct OOD prototypes. For any non-empty cache, we perform BIRCH clustering~\cite{birch} on the stored image features to identify different OOD patterns associated with the same ID class. Samples within the same cluster are aggregated by averaging to form a single OOD prototype. Similarly, ID training samples are aggregated by class to obtain one ID prototype per class. We denote the prototypes as
$P=[\mathbf{p}_1,\mathbf{p}_2,...,\mathbf{p}_C,\mathbf{p}_{C+1},...,\mathbf{p}_{C+M}]$,
 where $\mathbf{p}_1,...,\mathbf{p}_C$ are ID prototypes, and $\mathbf{p}_{C+1},...,\mathbf{p}_{C+M}$ are constructed OOD prototypes. $C$ is the number of ID classes, and $M$ is the number of constructed OOD prototypes.

Given a test image feature $\mathbf{v}=\mathcal{F}(\mathbf{x})\in\mathbb{R}^{D}$, we use the NegLabel scoring function (Eq.~\ref{neglabel_score}) to measure whether the sample belongs to ID. Inspired by FA~\cite{FA}, we find that multiplying the OOD prototype term in the denominator of $S_{\text{nl}}$ by an integer coefficient $K$ further improves the model’s OOD detection performance. More analyses about $K$ are provided in Section~\ref{sec:analysis}. Thus, our final scoring function is defined as:
\begingroup
\small
\begin{equation}
\label{s_proto}
S_{\text{DynProto}}(\mathbf{v}) = 
\frac{
    \sum_{i=1}^{C} e^{\cos(\mathbf{v}, \mathbf{p}_i) / \tau}
}{
    \sum_{i=1}^{C} e^{\cos(\mathbf{v}, \mathbf{p}_i) / \tau}
    + K\,\sum_{j=C+1}^{C+M} e^{\cos(\mathbf{v}, \mathbf{p}_j) / \tau}
}.
\end{equation}
\endgroup

\subsection{Cold Start}
\label{sec:cold_start}
During the first $T_{\text{cold}}$ iterations of testing, when the caches contain only a few samples, we use the MSP score~\cite{msp} as the base OOD detector for unimodal models, while for VLMs we adopt the MCM score~\cite{mcm} to determine which test samples should be cached. We also experiment with other methods as base detectors, and detailed analyses are provided in Section~\ref{sec:results}. The score threshold for sample collection is denoted by $\theta$. Since the score distribution of OOD samples is unavailable at test time, $\theta$ is estimated as the $\beta$-th percentile of the OOD score distribution derived from ID training data, with $\beta = 5$ in our experiments.

As testing proceeds and more samples are accumulated, we stop using base detector and instead use DynProto score for selecting potential OOD samples, as DynProto score begins to outperform the base detector. Inspired by OWTTT~\cite{OWTTT}, a dynamic threshold $\alpha$ is introduced to enable the model to adapt to varying sample distributions across different batches. As reported in OWTTT, the OOD scores exhibit a bimodal distribution, and the adaptive threshold is obtained by minimizing intra-class variance. Accordingly, the coefficient $\alpha$ is computed as follows in OWTTT:
\begingroup
\small
\begin{equation}
\label{adaptive_threshold}
\begin{aligned}
\min_{\alpha} \;
\frac{1}{N_{\text{id}}} &\sum_{i=1}^{N_{\text{b}}}
\left[ S(x_i) - \frac{1}{N_{\text{id}}} 
\sum_{j=1}^{N_{\text{b}}} \indicator{S(x_j) > \alpha} S(x_j) \right]^2 \\
+
\frac{1}{N_{\text{ood}}} &\sum_{i=1}^{N_{\text{b}}}
\left[ S(x_i) - \frac{1}{N_{\text{ood}}} 
\sum_{j=1}^{N_{\text{b}}} \indicator{S(x_j) \le \alpha} S(x_j) \right]^2,
\end{aligned}
\end{equation}
\endgroup
where $N_{\text{id}} = \sum_{i}^{N_\text{b}} \mathbf{1}_{\{ S(x_i) > \alpha \}}$, $N_{\text{ood}} = \sum_{i}^{N_\text{b}} \mathbf{1}_{\{ S(x_i) \leq \alpha \}}$ and $N_b$ is batch size. We search $\alpha$ over the interval $(0,1)$ and select the optimal value according to Eq.~(\ref{adaptive_threshold}). Samples with OOD scores lower than $\alpha$ are stored in the cache associated with their predicted ID class.

Finally, after the COPC module and FOPR module, we determine whether a test sample is ID or OOD by computing the DynProto score (Eq.~\ref{s_proto}). The overall framework of our proposed method is summarized in Algorithm.~\ref{alg:DynProto}.

\begin{algorithm}[!t]
\small
\caption{DynProto}
\label{alg:DynProto}
\begin{algorithmic}[1]
\REQUIRE Test sample $\mathbf{x}$, model $f$, cache $\mathcal{C}$, thresholds $\theta,\alpha$, max cache size $m$, cold-start round $T_{\text{cold}}$, batch index $t$
\STATE $\mathbf{v} \leftarrow f(\mathbf{x})$
\IF{$\mathcal{C}\neq\varnothing$ and $t\ge T_{\text{cold}}$}
    \STATE $S_{\text{DynProto}}\!=\!\frac{\sum_{i=1}^{C} e^{\cos(\mathbf{v},\mathbf{p}_i)/\tau}}
    {\sum_{i=1}^{C} e^{\cos(\mathbf{v},\mathbf{p}_i)/\tau}+K\sum_{j=C+1}^{C+M} e^{\cos(\mathbf{v},\mathbf{p}_j)/\tau}}$
    \STATE Compute $\alpha$ according to Eq.~(\ref{adaptive_threshold})
\ELSE
    \STATE Obtain $S_{\text{base}}$ from the base detector
\ENDIF
\IF{$(t<T_{\text{cold}}\!\wedge\! S_{\text{base}}<\theta)$ \textbf{or} $(t\ge T_{\text{cold}}\!\wedge\! S_{\text{DynProto}}<\alpha)$}
    \STATE Update $\mathcal{C}[\hat{y}]$ with $\mathbf{v}$
\ENDIF
\IF{$\mathcal{C}\neq\varnothing$}
    \STATE Compute ID prototypes $\mathbf{p}_c=\frac{1}{N_c}\!\sum_i\!\mathbf{v}_i^c$
    \STATE Cluster each $\mathcal{C}_c$ via BIRCH and aggregate them into OOD prototypes 
$\{\mathbf{p}_{C+1}, \mathbf{p}_{C+2}, \ldots, \mathbf{p}_{C+M}\}$
    \STATE $S\leftarrow S_{\text{DynProto}}$
\ELSE
    \STATE $S\leftarrow S_{\text{base}}$
\ENDIF
\RETURN $S$
\end{algorithmic}
\end{algorithm}

\begin{table*}[!ht]
\centering
\small
\caption{Comparison with SOTA methods on the CIFAR100 OOD benchmarks. The ``External Text'' column denotes whether the method uses external textual information. The best results are in bold, and the second best are underlined.}
\vspace{-1em}
\resizebox{\textwidth}{!}{
\begin{tabular}{cccccccccccccccc}
\toprule
\multirow{2}{*}{\textbf{Method}} 
& \multirow{2}{*}{\begin{tabular}[c]{@{}c@{}}\textbf{External}\\ \textbf{Text}\end{tabular}} 
& \multicolumn{2}{c}{SVHN} & \multicolumn{2}{c}{LSUN-R} & \multicolumn{2}{c}{LSUN-C} & \multicolumn{2}{c}{iSUN} & \multicolumn{2}{c}{Textures} & \multicolumn{2}{c}{Places365}  
& \multicolumn{2}{c}{\textbf{Average}} \\ 
\cmidrule(lr){3-4} \cmidrule(lr){5-6} \cmidrule(lr){7-8} \cmidrule(lr){9-10} \cmidrule(lr){11-12} \cmidrule(lr){13-14} \cmidrule(lr){15-16}
& & \textbf{FPR95}$\downarrow$ & \textbf{AUROC}$\uparrow$
& \textbf{FPR95}$\downarrow$ & \textbf{AUROC}$\uparrow$
& \textbf{FPR95}$\downarrow$ & \textbf{AUROC}$\uparrow$
& \textbf{FPR95}$\downarrow$ & \textbf{AUROC}$\uparrow$
& \textbf{FPR95}$\downarrow$ & \textbf{AUROC}$\uparrow$
& \textbf{FPR95}$\downarrow$ & \textbf{AUROC}$\uparrow$ 
& \textbf{FPR95}$\downarrow$ & \textbf{AUROC}$\uparrow$ \\ 
\midrule

\multicolumn{16}{c}{\textbf{Vision Model based Methods}} \\
\midrule
MSP~\cite{msp} & – & 82.02 & 75.19 & 87.28 & 67.13 & 76.44 & 78.63 & 88.00 & 68.49 & 85.19 & 71.20 & 85.28 & 70.84 & 84.04 & 71.91 \\
Energy~\cite{energy} & – & 88.03 & 81.30 & 75.17 & 77.77 & 58.19 & 88.11 & 78.61 & 76.79 & 85.00 & 70.99 & 79.95 & 76.21 & 77.49 & 78.53 \\
ReAct~\cite{react} & – & 96.75 & 69.13 & 68.03 & 86.44 & 77.21 & 78.84 & 74.78 & 82.86 & 92.07 & 67.15 & 89.72 & 59.99 & 83.09 & 74.07 \\
DICE~\cite{dice} & – & 60.06 & 88.18 & 55.03 & 88.23 & 36.40 & 92.98 & 52.49 & 88.50 & 61.27 & 77.22 & 73.89 & \underline{81.18} & 56.52 & 86.05 \\
ASH-S~\cite{ash} & – & 24.75 & 95.79 & 54.06 & 89.54 & 29.98 & 94.14 & 48.15 & 90.93 & 34.60 & 92.11 & 76.96 & 79.22 & 44.75 & 90.29 \\
OptFS~\cite{optfs} & – & 73.61 & 84.96 & 69.52 & 83.61 & 47.98 & 90.01 & 70.56 & 84.39 & 61.64 & 85.63 & 80.96 & 74.37 & 67.38 & 83.83 \\
SLE~\cite{sle} & – & \underline{2.90} & \underline{99.37} & \underline{5.17} & \underline{98.81} & \textbf{2.42} & \textbf{99.34} & \underline{9.25} & \underline{97.74} & 80.09 & 51.91 & \underline{59.44} & 77.38 & \underline{26.55} & 87.43 \\
CADRef~\cite{cadref} & – & 18.28 & 96.69 & 47.45 & 90.26 & 27.22 & 94.70 & 42.10 & 91.59 & 28.72 & 94.13 & 78.30 & 75.91 & 40.34 & 90.55 \\
VIM~\cite{vim} & – & 35.05 & 93.57 & 24.65 & 95.50 & 40.06 & 92.76 & 23.22 & 95.63 & \underline{19.75} & \underline{95.89} & 83.89 & 75.61 & 37.77 & \underline{91.49} \\
\rowcolor{red!15}
\textbf{Ours} & – & \textbf{0.22} & \textbf{99.94} & \textbf{0.39} & \textbf{99.90} & \underline{14.44} & \underline{97.56} & \textbf{0.99} & \textbf{99.75} & \textbf{16.81} & \textbf{96.34} & \textbf{53.83} & \textbf{87.38} & \textbf{14.45} & \textbf{96.81} \\

\midrule
\multicolumn{16}{c}{\textbf{Vision Language Model based Methods}} \\
\midrule
MCM~\cite{mcm} & \xmark & 83.53 & 87.34 & 83.07 & 82.26 & 71.47 & 87.49 & 80.90 & 81.93 & 98.01 & 66.71 & 99.68 & 52.78 & 86.11 & 76.42 \\
CoOp~\cite{coop} & \xmark & 20.55 & 96.43 & 65.98 & 83.60 & 45.04 & 89.48 & 70.37 & 80.61 & 74.04 & 77.70 & 94.09 & 58.65 & 61.68 & 81.09 \\
LoCoOp~\cite{locoop} & \xmark & 15.67 & 97.28 & 57.01 & 87.75 & 38.30 & 92.49 & 61.65 & 86.77 & 78.01 & 77.55 & 86.11 & 71.10 & 56.12 & 85.49 \\
FA~\cite{FA} & \xmark & 19.08 & 96.82 & 55.24 & 87.86 & 44.07 & 90.80 & 60.60 & 86.25 & 24.32 & \underline{94.97} & \underline{48.85} & \underline{88.68} & 42.03 & 90.89 \\
AdaND~\cite{adand} & \xmark & \underline{1.17} & \underline{99.67} & \underline{12.05} & \underline{96.75} & \underline{5.63} & \underline{98.35} & \underline{18.77} & \underline{95.68} & \underline{22.71} & 92.48 & 65.96 & 71.98 & \underline{21.05} & \underline{92.49} \\
NegLabel~\cite{neglabel} & \cmark & 81.20 & 70.14 & 68.53 & 88.11 & 70.43 & 87.18 & 67.84 & 88.08 & 73.39 & 76.52 & 97.02 & 48.68 & 76.40 & 76.45 \\
CSP~\cite{csp} & \cmark & 99.40 & 70.95 & 66.43 & 89.81 & 78.82 & 85.97 & 66.39 & 89.57 & 26.04 & 92.59 & 88.99 & 67.22 & 71.01 & 82.69 \\
NegRefine~\cite{negrefine} & \cmark & 55.51 & 88.06 & 74.24 & 85.41 & 52.57 & 90.64 & 74.73 & 84.97 & 25.25 & 92.70 & 84.89 & 64.18 & 61.20 & 84.33 \\
MCM+DCAC~\cite{dcac} & \cmark & 31.80 & 94.79 & 66.96 & 86.61 & 48.42 & 91.25 & 68.85 & 85.80 & 97.79 & 66.86 & 99.66 & 53.27 & 68.91 & 79.76 \\

\rowcolor{red!15}
\textbf{Ours} & \xmark & \textbf{0.03} & \textbf{99.97} & \textbf{4.01} & \textbf{98.97} & \textbf{1.76} & \textbf{99.34} & \textbf{4.73} & \textbf{98.71} & \textbf{18.58} & \textbf{95.82} & \textbf{12.22} & \textbf{93.90} & \textbf{6.89} & \textbf{97.79} \\

\bottomrule
\end{tabular}
}
\vspace{-0.5em}
\label{tab:cifar100_far}
\end{table*}

\begin{table*}[!ht] \centering \small \caption{Comparison with SOTA methods on the ImageNet OOD benchmarks.} 
\vspace{-1em}
\resizebox{\textwidth}{!}{ \begin{tabular}{cccccccccccccccc} \toprule \multirow{2}{*}{\textbf{Method}} & \multirow{2}{*}{\begin{tabular}[c]{@{}c@{}}\textbf{External}\\ \textbf{Text}\end{tabular}} & \multicolumn{2}{c}{SUN} & \multicolumn{2}{c}{Textures} & \multicolumn{2}{c}{iNaturalist} & \multicolumn{2}{c}{Places} & \multicolumn{2}{c}{NINCO} & \multicolumn{2}{c}{SSB-hard} & \multicolumn{2}{c}{\textbf{Average}} \\ \cmidrule(lr){3-4} \cmidrule(lr){5-6} \cmidrule(lr){7-8} \cmidrule(lr){9-10} \cmidrule(lr){11-12} \cmidrule(lr){13-14} \cmidrule(lr){15-16} & & \textbf{FPR95}$\downarrow$ & \textbf{AUROC}$\uparrow$ & \textbf{FPR95}$\downarrow$ & \textbf{AUROC}$\uparrow$ & \textbf{FPR95}$\downarrow$ & \textbf{AUROC}$\uparrow$ & \textbf{FPR95}$\downarrow$ & \textbf{AUROC}$\uparrow$ & \textbf{FPR95}$\downarrow$ & \textbf{AUROC}$\uparrow$ & \textbf{FPR95}$\downarrow$ & \textbf{AUROC}$\uparrow$ & \textbf{FPR95}$\downarrow$ & \textbf{AUROC}$\uparrow$ \\ \midrule \multicolumn{16}{c}{\textbf{Vision Model based Methods}} \\ \midrule MSP~\cite{msp} & – & 68.58 & 81.75 & 66.15 & 80.46 & 52.73 & 88.42 & 71.59 & 80.63 & 75.94 & 79.97 & 84.54 & 72.16 & 69.92 & 80.57 \\ Energy~\cite{energy} & – & 58.28 & 86.73 & 52.30 & 86.73 & 53.96 & 90.59 & 65.43 & 84.12 & 77.63 & 79.69 & 83.87 & 72.35 & 65.25 & 83.37 \\ ReAct~\cite{react} & – & 24.01 & 94.41 & 45.83 & 90.45 & 19.55 & 96.39 & 33.45 & 91.93 & 71.21 & 80.15 & 79.07 & 72.81 & 45.52 & 87.69 \\ ASH-S~\cite{ash} & – & 27.96 & 94.02 & \textbf{11.97} & \textbf{97.60} & 11.49 & 97.87 & 39.83 & 90.98 & 65.02 & 82.77 & 82.53 & 70.49 & 39.80 & 88.96 \\ OptFS~\cite{optfs} & – & 35.31 & 93.13 & 23.08 & 95.74 & 16.79 & 96.88 & 44.78 & 90.42 & 72.69 & 80.80 & 85.77 & 69.73 & 46.40 & 87.78 \\ 
SLE~\cite{sle} & – & 44.82 & 81.26 & 39.89 & 84.12 & 17.26 & 96.72 & \textbf{21.53} & \underline{92.84} & 69.13 & 70.45 & \underline{64.01} & 76.50 & 42.77 & 83.65 \\
CADRef~\cite{cadref} & – & 39.23 & 91.26 & \underline{12.60} & \underline{97.14} & 16.08 & 96.90 & 51.12 & 87.80 & 64.89 & \underline{85.36} & 78.79 & 74.58 & 43.79 & 88.84 \\ 
OODD~\cite{oodd} & – & 42.36 & 92.07 & 16.16 & 97.01 & \underline{7.13} & \underline{98.71} & 53.59 & 87.13 & \underline{62.48} & 79.41 & 80.35 & 65.90 & 43.67 & 86.71 \\
LINe~\cite{line} & – & \underline{19.48} & \underline{95.26} & 22.54 & 94.44 & 12.26 & 97.56 & \underline{28.52} & \textbf{92.85} & 66.62 & 81.13 & 75.78 & 77.14 & \underline{37.53} & \underline{89.73} \\
MSP+DCAC~\cite{dcac} & – & 36.68 & 91.46 & 50.83 & 85.99 & 24.84 & 95.09 & 46.78 & 88.51 & 70.58 & 82.61 & 74.49 & \underline{77.38} & 50.70 & 86.84 \\
\rowcolor{red!15} \textbf{Ours} & – & \textbf{17.77} & \textbf{96.15} & 23.46 & 93.80 & \textbf{2.14} & \textbf{99.48} & 31.53 & 91.90 & \textbf{46.72} & \textbf{87.36} & \textbf{49.50} & \textbf{86.25} & \textbf{28.52} & \textbf{92.49} \\ \midrule \multicolumn{16}{c}{\textbf{Vision Language Model based Methods}} \\ \midrule MCM~\cite{mcm} & \xmark & 37.22 & 92.55 & 58.35 & 85.83 & 31.95 & 94.16 & 42.98 & 90.10 & 79.50 & 74.57 & 89.90 & 62.95 & 56.65 & 83.36 \\ LoCoOp~\cite{locoop} & \xmark & 23.44 & 95.07 & 42.28 & 90.19 & 16.05 & 96.86 & 32.87 & 91.98 & 75.16 & 72.97 & 85.61 & 68.22 & 45.90 & 85.88 \\ AdaND~\cite{adand} & \xmark & 17.08 & 95.86 & 21.76 & 93.01 & 4.19 & 98.91 & 20.95 & 94.55 & 80.35 & 69.69 & 81.22 & 66.64 & 37.59 & 86.44 \\ FA~\cite{FA} & \xmark & 27.65 & 93.46 & 29.50 & 92.93 & 14.49 & 96.48 & 31.09 & 92.44 & 65.61 & 79.01 & \underline{70.83} & \underline{78.21} & 39.86 & 88.76 \\ 
NegLabel~\cite{neglabel} & \cmark & 20.53 & 95.49 & 43.56 & 90.22 & 1.91 & 99.49 & 35.59 & 91.64 & 69.99 & 76.13 & 85.68 & 69.39 & 42.88 & 87.06 \\ NegRefine~\cite{negrefine} & \cmark & 22.93 & 94.64 & \underline{21.15} & 94.69 & 1.51 & 99.57 & 39.10 & 90.42 & 63.93 & \underline{81.00} & 78.79 & 74.59 & 37.90 & 89.15 \\ CSP~\cite{csp} & \cmark & 13.66 & 96.66 & 25.52 & 93.86 & 1.54 & 99.60 & 29.32 & 92.90 & 67.14 & 79.13 & 78.88 & 73.61 & 36.01 & 89.29 \\ AdaNeg~\cite{adaneg} & \cmark & \underline{9.50} & \underline{97.44} & 31.27 & 94.93 & \textbf{0.59} & \textbf{99.71} & 34.34 & 94.55 & 60.10 & 78.30 & 74.91 & 75.11 & 35.12 & 90.01 \\ 
SynOOD~\cite{synood} & \cmark & 21.63 & 95.83 & 25.24 & \underline{95.08} & 1.97 & 99.43 & \textbf{10.27} & \textbf{97.80} & \underline{57.12} & 76.28 & 85.94 & 69.58 & \underline{33.70} & \underline{89.00} \\
MCM+DCAC~\cite{dcac} & \xmark & 17.56 & 96.65 & 42.43 & 90.72 & 7.39 & 98.27 & 29.62 & 93.05 & 75.92 & 75.02 & 84.05 & 67.85 & 42.83 & 86.98 \\
\rowcolor{red!15} \textbf{Ours} & \xmark & \textbf{7.37} & \textbf{98.55} & \textbf{10.94} & \textbf{97.48} & \underline{1.22} & \underline{99.64} & \underline{18.53} & \underline{95.45} & \textbf{51.04} & \textbf{84.28} & \textbf{37.51} & \textbf{86.78} & \textbf{21.10} & \textbf{93.70} \\ \bottomrule \end{tabular} }
\vspace{-1em} 
\label{tab:imagenet} \end{table*}

\section{Experiments}
\subsection{Experimental Setup}
\noindent\textbf{Datasets.} We perform extensive experiments using both the CIFAR-100~\cite{cifar} and ImageNet~\cite{imagenet} the  as ID datasets. In line with prior work~\cite{mcm,cadref}, we select SVHN~\cite{SVHN}, LSUN-R~\cite{LSUN}, LSUN-C, iSUN~\cite{iSUN}, Texture~\cite{dtd}, and Places365~\cite{places} as the OOD datasets for CIFAR-100. For ImageNet, we use iNaturalist~\cite{inaturalist}, SUN~\cite{sun}, Places~\cite{places}, and Textures~\cite{dtd} as far-OOD datasets, while NINCO~\cite{ninco} and SSB-hard~\cite{ssb_hard} are used as near-OOD datasets. We also verify the effectiveness of DynProto on two cross-domain ID datasets: EuroSAT~\cite{eurosat} and BIMCV-COVID19+~\cite{bimcv}(Sectiom~\ref{cross_domain}).

\noindent\textbf{Implementation Details.} 
We implement our method in PyTorch and conduct all experiments on a NVIDIA GeForce RTX 3090 GPU with a batch size of 512. Following the test-time adaptation protocol, we mix and randomly shuffle ID and OOD samples. Each experiment is performed using five distinct random seeds for shuffling. The cache size is set to $m=30$ for each class. The threshold $\theta$ is defined as the $\beta=5$th percentile of the OOD score distribution on the ID training set. The cold-start period $T_{\text{cold}}$ is set to 5, and the coefficient $K$ is set to 5.

\noindent\textbf{Metric.} 
We use two evaluation metrics: the false positve rate at 95\% true positive rate (FPR95) and the area under the receiver operating characteristic curve (AUROC).

\noindent\textbf{Baselines.}
We adopt a comprehensive set of competitive baselines for comparison. For Vision-Model-based OOD detection, we follow standard practice and use DenseNet-101~\cite{densenet} and ResNet-50~\cite{resnet} as backbones for CIFAR-100 and ImageNet, respectively. The compared methods include MSP~\cite{msp}, Energy~\cite{energy}, ReAct~\cite{react}, DICE~\cite{dice}, VIM~\cite{vim}, ASH-S~\cite{ash}, LINe~\cite{line}, OptFs~\cite{optfs}, OODD~\cite{oodd}, SLE~\cite{sle}, CADRef~\cite{cadref} and DCAC~\cite{dcac}.
For VLM-based OOD detection, we use CLIP-B/16 as the backbone and compare against MCM~\cite{mcm}, CoOp~\cite{coop}, LoCoOp~\cite{locoop}, NegLabel~\cite{neglabel}, CSP~\cite{csp}, AdaNeg~\cite{adaneg}, AdaND~\cite{adand}, NegRefine~\cite{negrefine}, FA~\cite{FA}, DCAC and SynOOD~\cite{synood} as baselines. Among them, OODD, AdaNeg, AdaND and DCAC are test-time adaptation methods. All baselines are evaluated using their original hyperparameter configurations to ensure fair comparison.

\subsection{Experimental Results}
\label{sec:results}

\noindent\textbf{Performance on CIFAR100 benchmarks.} 
Table~\ref{tab:cifar100_far} summarizes the results on the CIFAR-100 OOD benchmark. DynProto consistently achieves state-of-the-art performance with both the DenseNet101 and CLIP-B/16 backbones. For DenseNet101, DynProto obtains an average FPR95 of 14.45\% and an AUROC of 96.81\%, improving upon the strongest competing results by 12.10\% and 5.32\%, respectively. In particular, it achieves nearly perfect detection on SVHN, LSUN-R, and iSUN, while also yielding substantial improvements on the more challenging Textures and Places365 datasets. With CLIP-B/16, DynProto further reduces the average FPR95 to 6.89\% and increases the AUROC to 97.79\%, outperforming the best baseline by 14.16\% and 5.30\%, respectively, and ranking first across all six OOD datasets. Notably, methods relying on external textual information perform poorly on SVHN because digit images have limited semantic correspondence with the predefined textual OOD labels. In contrast, DynProto does not depend on manually designed OOD concepts; instead, it adaptively captures the feature distribution of test-time digit samples and organizes them into representative OOD prototypes. These prototypes allow easily detected OOD samples to provide additional evidence for identifying similar but more ambiguous samples, resulting in near-zero FPR95 on SVHN and robust performance across diverse distribution shifts.

\noindent\textbf{Performance on ImageNet benchmarks.}
Table~\ref{tab:imagenet} reports the results on the ImageNet OOD benchmark. DynProto achieves the best average performance on both ResNet-50 and CLIP-B/16. On ResNet-50, it reduces the average FPR95 by 9.03\% and improves AUROC by 2.76\% over the strongest baselines, with particularly strong results on iNaturalist, NINCO, and SSB-hard. On CLIP-B/16, DynProto further outperforms the second-best results by 12.60\% in FPR95 and 3.69\% in AUROC. The improvement is especially pronounced on near-OOD datasets: on SSB-hard, DynProto reduces FPR95 by 33.32\% and improves AUROC by 8.57\%. These results indicate that the dynamically constructed prototypes effectively capture both obvious and fine-grained OOD patterns, enabling more reliable detection of samples that are visually or semantically close to ID classes.

\begin{figure}[h]
    \centering
    \includegraphics[width=0.45\textwidth]{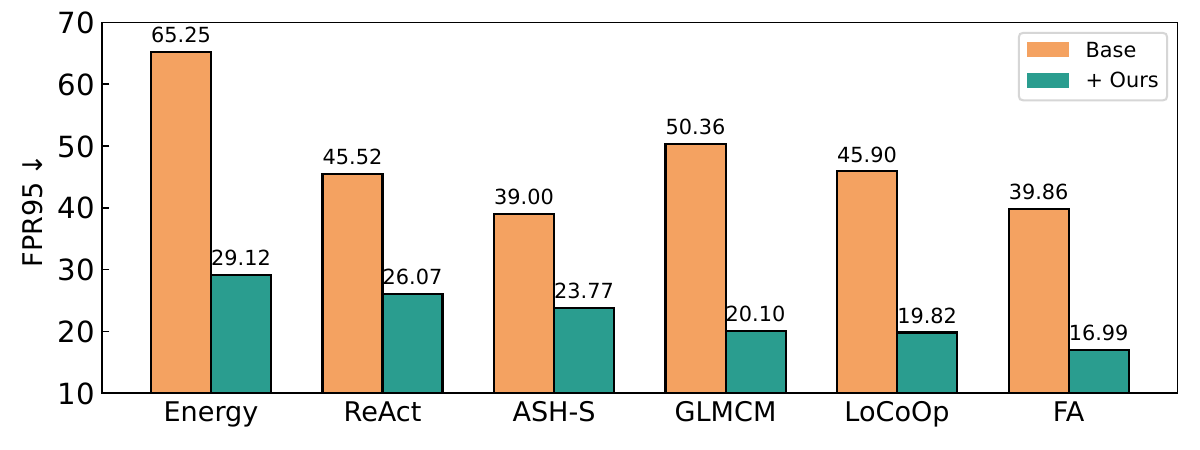}
    \vspace{-1.0em}
    \caption{FPR95 of DynProto integrated with various baselines on ImageNet OOD benchmark. }
    \label{fig:more_baseline}
\end{figure}

\noindent\textbf{Performance under Various Base Detectors.}  
In DynProto, we employ the basic MSP~\cite{msp} and MCM~\cite{mcm} scores as our base detectors. As illustrated in Figure~\ref{fig:more_baseline}, we also integrate other baseline methods as base detectors for comparison. Among them, Energy~\cite{energy}, ReAct~\cite{react}, and ASH-S~\cite{ash} are unimodal approaches, while GLMCM~\cite{glmcm}, LoCoOp~\cite{locoop}, and FA~\cite{FA} are VLM-based methods. The figure reveals a clear trend: when the base detector exhibits stronger performance, DynProto can capture more OOD patterns during the cold-start phase. This provides a more reliable foundation for subsequent process, ultimately leading to superior overall performance.

\noindent\textbf{Performance under Various Backbones.}  
We also evaluate our method across different backbone architectures to verify its generalizability. As illustrated in Table~\ref{tab:different_backbone}, our method consistently outperforms all compared approaches regardless of the underlying architecture. Specifically, compared with the strongest baseline, it achieves significant reductions in average FPR95 by 23.06, 35.25, and 32.24 points on ViT-B/16~\cite{vit}, Swin-B~\cite{swin}, and ConvNeXt-B~\cite{convnet}, respectively. Meanwhile, the corresponding AUROC improvements are 5.72, 6.23, and 5.27 points. These results demonstrate both the effectiveness and the architecture-agnostic nature of DynProto.

\begin{table}[t]
\centering
\caption{Average Performance over ImageNet far-OOD benchmark
using different backbones.}
\label{tab:different_backbone}

\setlength{\tabcolsep}{2.5pt}
\renewcommand{\arraystretch}{0.95}

\resizebox{\columnwidth}{!}{
\begin{tabular}{@{}lcccccc@{}}
\toprule
\multirow{2}{*}{Method}
& \multicolumn{2}{c}{ViT-B/16}
& \multicolumn{2}{c}{Swin-B}
& \multicolumn{2}{c}{ConvNeXt-B} \\
\cmidrule(lr){2-3}
\cmidrule(lr){4-5}
\cmidrule(lr){6-7}
& FPR95$\downarrow$ & AUROC$\uparrow$
& FPR95$\downarrow$ & AUROC$\uparrow$
& FPR95$\downarrow$ & AUROC$\uparrow$ \\
\midrule

MSP~\cite{msp}
& 61.83 & 83.10
& 63.05 & 81.67
& 60.13 & 80.29 \\

Energy~\cite{energy}
& 67.39 & 74.27
& 73.99 & 65.93
& 91.18 & 53.21 \\

ReAct~\cite{react}
& 67.10 & 81.76
& 64.25 & 83.78
& 74.45 & 79.99 \\

DICE~\cite{dice}
& 90.50 & 70.48
& 86.81 & 43.17
& 89.67 & 42.02 \\

ASH-S~\cite{ash}
& 99.64 & 16.46
& 99.35 & 17.66
& 98.67 & 15.66 \\

OptFS~\cite{optfs}
& 61.74 & 85.82
& 63.30 & 86.15
& 56.51 & 87.34 \\

VIM~\cite{vim}
& 43.70 & 87.89
& 63.03 & 83.91
& 53.65 & 86.90 \\

CADRef~\cite{cadref}
& 55.50 & 87.95
& 54.05 & 87.88
& 51.88 & 88.88 \\

\rowcolor{red!12}
\textbf{Ours}
& \textbf{20.64} & \textbf{93.67}
& \textbf{18.80} & \textbf{94.11}
& \textbf{19.64} & \textbf{94.15} \\

\bottomrule
\end{tabular}
}
\vspace{-2.0em}
\end{table}

\subsection{Analysis of DynProto}
\label{sec:analysis}

To ensure consistency, all subsequent analyses are conducted on ImageNet OOD benchmark using CLIP-B/16.

\noindent\textbf{Ablation Study.} 
The ablation study primarily evaluates the effectiveness of the COPC module and the FOPR module in DynProto. Without the two modules, performance matches the base detector (BD). When only the COPC module is applied, each cached sample is directly treated as an OOD prototype for scoring. As shown in Table~\ref{tab:ablation}, the COPC module effectively gathers diverse OOD patterns during testing, improving the model’s ability to approximate the OOD samples distribution. The FOPR module further clusters and aggregates these patterns, leading to more representative and stable OOD prototypes.

\begin{table}[h]
\centering
\caption{Ablation study of DynProto.}
\resizebox{\linewidth}{!}{%
\begin{tabular}{ c c c c c c c}
\toprule
\multirow{2}{*}{BD} & 
\multirow{2}{*}{COPC} & 
\multirow{2}{*}{FOPR} & 
\multicolumn{2}{c}{far-OOD} & 
\multicolumn{2}{c}{near-OOD} \\
\cmidrule(lr){4-5} \cmidrule(lr){6-7}
& & & FPR95$\downarrow$ & AUROC$\uparrow$ & FPR95$\downarrow$ & AUROC$\uparrow$ \\
\midrule
\cmark  &               &               &  
42.63 & 90.66   & 84.70 & 68.76 \\
\cmark  & \cmark    &               &  
9.78  & 97.71   & 48.12 & 84.67 \\
\cmark  & \cmark    & \cmark    &  
\textbf{9.52} & \textbf{97.78} & \textbf{44.28} & \textbf{85.53} \\
\bottomrule
\end{tabular}%
}
\label{tab:ablation}
\end{table}

\noindent\textbf{Impact of Cache Initialization.}
Before testing begins, we initialize the cache for each class as empty. Following OODD~\cite{oodd}, we further explore several alternative cache initialization strategies and analyze their effects on the final performance: (1) C-Out: randomly crop ID training samples and initialize the cache using cropped views with lower OOD scores; 
(2) T-Out: randomly fill the cache with OOD samples whose categories differ from the real OOD types; (3) D-Out: initialize the cache using a subset of real OOD samples. For each cache initialization strategy, we pre-store the features of 800 samples into the cache before evaluation. As shown in Table~\ref{tab:initialized}, the C-Out strategy yields the worst performance. This is because, during the cropping process, some ID-related information is inevitably stored in the cache, leading to cache contamination. The T-Out strategy achieves the best results, as initializing the cache with real OOD samples effectively mitigates the transient instability during the cold-start phase. However, in practical testing scenarios, it is often infeasible to obtain real OOD samples in advance, making this initialization strategy unrealistic. The performance of D-Out is comparable to that of initializing the cache as empty. Therefore, from a cost-efficiency perspective, we initialize all caches as empty.

\begin{table}[ht]

\centering
\small

\caption{Performance with different cache initialized strategies.}
\begin{tabular}{@{} c c c c c @{}}
\toprule
Strategy & C-Out & D-Out & Empty & T-Out \\
\midrule
FPR95$\downarrow$ & 22.53 & 21.77 & 21.10 & \textbf{16.88} \\
AUROC$\uparrow$  & 92.47 & 93.56 & 93.70 & \textbf{95.04}  \\
\bottomrule
\end{tabular}
\label{tab:initialized}
\end{table}

\noindent\textbf{Impact of Cluster Strategies.} 
In DynProto, the Fine-grained OOD Pattern Refinement module applies BIRCH to cluster the samples in each cache into refined OOD prototypes. We also evaluate K-means, VB-GMM, and a simple alternative that aggregates all samples in the same cache as a single prototype (AP). As shown in Table~\ref{tab:cluster}, VB-GMM yields the best performance but is about twice as slow as BIRCH. K-means requires manually specifying the number of clusters, and AP ignores intra-cache variation. Balancing effectiveness and efficiency, we adopt BIRCH as our refinement strategy.

\begin{table}[t]
\centering
\small
\caption{Performance with different cluster strategies.}
\begin{tabular}{ccccc}
\toprule
Strategy & Kmeans & AP & BIRCH & VB-GMM \\
\midrule
FPR95$\downarrow$ & 22.37 & 23.06 & 21.10 & \textbf{20.19} \\
AUROC$\uparrow$  & 93.59 & 93.33 & 93.70 & \textbf{93.86}  \\
\bottomrule
\end{tabular}
\label{tab:cluster}
\end{table}

\begin{figure}[h]
    \centering
    \includegraphics[width=0.45\textwidth]{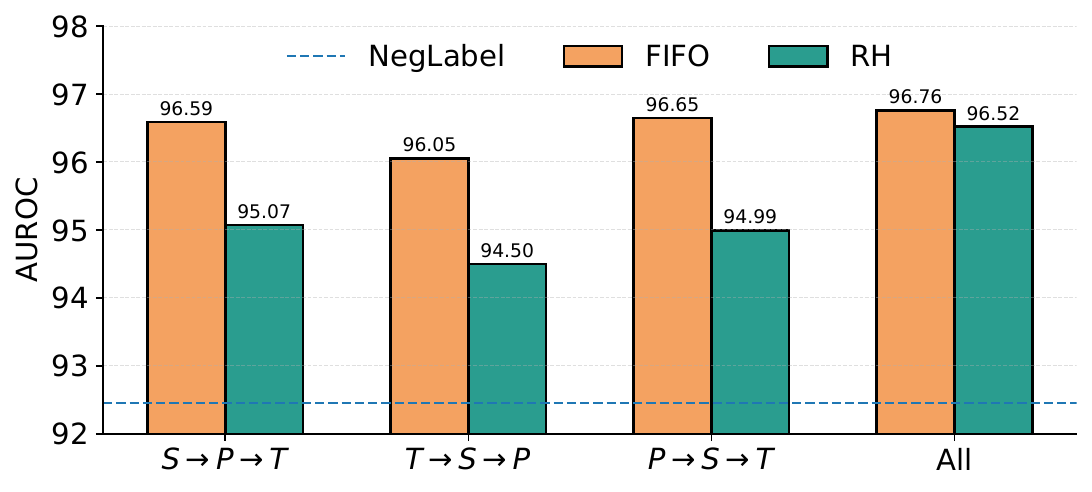}
    \vspace{-1em}
    \caption{AUROC for the temporal drift OOD scenario settings with different update strategies.  }
    \label{fig:temporal_drift}
\end{figure}

\noindent\textbf{Impact of Cache Update Strategies}
\label{Update_Strategies}
In our current implementation, we adopt a FIFO policy for cache updates, where the oldest entry is discarded once the cache is full and a new OOD candidate is added. Furthermore, following OODD~\cite{oodd}, we also evaluate an alternative strategy that replaces the cache entry with the highest OOD score when inserting new samples (RH). Considering that OOD types may exhibit  abrupt changes in real-world scenarios, we simulate a situation where different OOD datasets appear sequentially over time. For instance, if the order of OOD datasets is SUN $\rightarrow$ Places $\rightarrow$ Textures, it is denoted as $S \rightarrow P \rightarrow T$. We simulate four such sequences: $S \rightarrow P \rightarrow T$, $T\rightarrow S \rightarrow P$ and $P \rightarrow S \rightarrow T$. Additionally, we also test a scenario where all OOD datasets are mixed with the ID data and shuffled, denoted as ``All''. 
As shown in Fig~\ref{fig:temporal_drift},  our FIFO strategy consistently outperforms the RH-based cache update strategy under different temporal drift scenarios, and also surpasses the baseline method NegLabel~\cite{neglabel} that relies on external textual labels.  According to the principle  of temporal locality, the current test sample is more likely  to be correlated with its recent samples. Therefore, adopting the FIFO strategy allows the model to better capture  the evolving OOD in real time. In contrast, although the RH algorithm maintains a cache containing many representative OOD patterns, as testing progresses,  the cache becomes dominated by earlier OOD samples with relatively low scores that are rarely replaced. Consequently, new OOD patterns fail to enter the cache, leading to reduced cache circulation and degraded adaptability to distributional drift. Moreover, in realistic OOD detection scenarios, various OOD samples often appear in a mixed and unpredictable manner. Even under such conditions, our model consistently outperforms existing baselines. This demonstrates that our method can accurately capture the dynamic evolution of OOD distributions, maintaining strong robustness and adaptability in complex testing environments. 

\noindent\textbf{Performance under Imbalance Data Ratio.} To investigate the stability of the model under imbalanced ratios of ID and OOD samples, we construct test sets with varying ID–OOD proportions.
We use ImageNet as the ID dataset and Textures as the OOD dataset, constructing six ID-to-OOD ratios: 500:1, 100:1, 50:1, 1:1, 1:50, and 1:100.
For ID-dominant settings (500:1, 100:1, 50:1), all 50,000 ID test samples are used with 100, 500, and 1,000 randomly selected OOD samples. For OOD-dominant settings (1:1, 1:50, 1:100), 5,000 OOD samples are used with 5,000, 100, and 50 randomly drawn ID samples, respectively.
As shown in Table~\ref{tab:imbalance}, DynProto consistently outperforms NegLabel across a wide range of ID–OOD mixture ratios (from 100:1 to 1:100), demonstrating the robustness of our approach. However, DynProto shows a potential limitation when the proportion of OOD samples is extremely small (e.g., 500:1). As the OOD ratio decreases, the model’s performance gradually degrades—for instance, when the ID:OOD ratio changes from 1:100 to 500:1, the FPR95 increases from 4.10\% to 67.00\%. This degradation occurs because a lower OOD ratio reduces the diversity of OOD patterns that can be captured in the cache. To alleviate this issue, inspired by AdaND~\cite{adand}, we introduce a small number of Gaussian noise samples into the test data to increase the proportion of OOD samples. When the ratio between ID and OOD samples is extremely imbalanced, this strategy helps mitigate the performance degradation of DynProto. For example, when the ID:OOD ratio is 500:1, the FPR95 of DynProto decreases from 67\% to 39\%. These Gaussian noise samples have lower OOD scores, which help reduce the value of the dynamic threshold $\alpha$ during its calculation and prevents more ID samples from being stored in the cache when the number of ID samples significantly exceeds the OOD samples.

\begin{table}[!t]
\small
\centering
\caption{FPR95 with different ratios of ID and OOD samples.}
\vspace{-1em}
\resizebox{\linewidth}{!}{
\begin{tabular}{ccccccc}
\toprule
ID:OOD Ratio & 500:1 & 100:1 & 50:1 & 1:1 & 1:50 & 1:100  \\
\midrule
NegLabel & 44.43 & 43.80 & 44.10 & 42.10 & 54.28 & 58.76  \\
DynProto & 67.00 &29.20 & 15.90 & 5.82 & 4.74 & 4.10  \\
DynProto+noise & 39.00 & 25.42 & 14.03 & 6.12 & 4.97 & 4.34 \\
\bottomrule
\end{tabular}
}
\label{tab:imbalance}
\vspace{-1em}
\end{table}

\noindent\textbf{Generalization to Other Domains.}
\label{cross_domain}
 Besides natural images, we also evaluated DynProto on the ID satellite image dataset EuroSAT~\cite{eurosat} and medical image dataset BIMCV-COVID19+~\cite{bimcv}. According to FA~\cite{FA} and OpenOOD~\cite{openood}, the OOD datasets for EuroSAT are the same as the ImageNet far-OOD datasets, and the OOD datasets for BIMCV-COVID19+ are CT-SCAN and X-Ray-Bone. As shown in Table~\ref{tab:comparison_other_ID}, DynProto outperforms NegLabel consistently across these cross-domain datasets.

\begin{table}[ht]
\centering
\small
\caption{Performance on cross-domain datasets.}
\resizebox{\linewidth}{!}{
\begin{tabular}{cccccc}
\toprule
\multirow{2}{*}{\centering ID Dataset} & \multicolumn{2}{c}{NegLabel~\cite{neglabel}} & \multicolumn{2}{c}{DynProto} \\
\cmidrule(lr){2-3} \cmidrule(lr){4-5}
 & FPR95$\downarrow$ & AUROC$\uparrow$ & FPR95$\downarrow$ & AUROC$\uparrow$ \\
\midrule
EuroSat       & 64.89 & 74.46 & \textbf{18.28} & \textbf{95.33} \\
BIMCV-COVID19+ & 50.28 & 81.61 & \textbf{17.66} & \textbf{97.54} \\
\bottomrule
\end{tabular}
}
\label{tab:comparison_other_ID}
\end{table}

\begin{figure*}[!t]
\centering
\includegraphics[width=\linewidth]{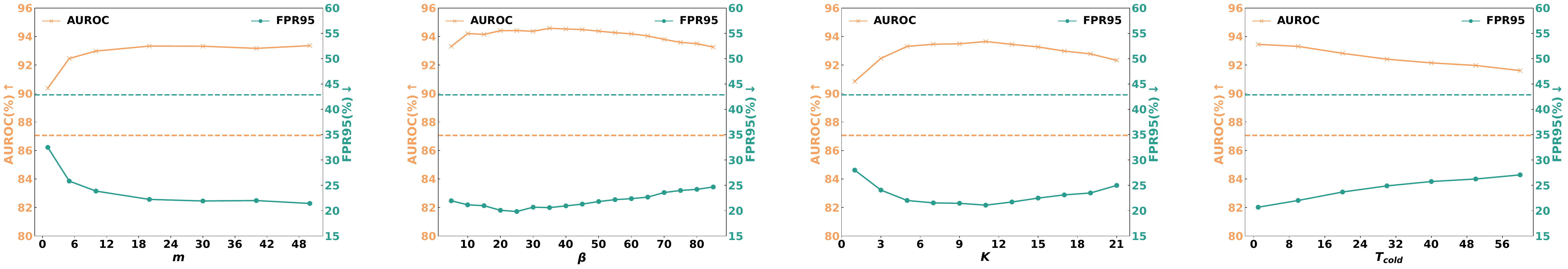}
\vspace{-2em}
\caption{Sensitivity analysis of hyper-parameters. Each dashed line represents performance of the baseline NegLabel~\cite{neglabel}.}

\label{fig:sensitivity}
\vspace{-1.5em}
\end{figure*}

\noindent\textbf{Hyperparameter Sensitivity Analysis.}  As shown in Figure~\ref{fig:sensitivity}, we conduct a series of hyperparameter sensitivity experiments on ImageNet OOD benchmark to evaluate how different hyperparameters influence the performance.  

We first analyze the impact of the cache capacity $m$. A larger cache consistently leads to better performance, as it enables the cache to store more diverse OOD patterns. 

Next, we investigate the effect of $\beta$, which determines the OOD score threshold for sample collection during the cold-start stage. In our experiments, $\beta$ ranges from 0 to 100, and we evaluate values between 5 and 85.  As illustrated, increasing $\beta$ initially enhances model performance, which subsequently declines when the parameters continue to grow. This trend can be explained as follows. When $\beta$ is small, only a limited number of OOD samples are admitted into the cache. As $\beta$ increases, more OOD samples are incorporated, improving the ability of DynProto to discriminate between ID and OOD samples. However, once $\beta$ exceed a threshold, the number of cached OOD samples reaches saturation, and additional increases cause more ID samples to be mistakenly stored,  causing cache contamination and a slight performance drop.

We then analyze the effect of the scaling factor $K$ in Eq.~(\ref{s_proto}) which can be formulated as $S(\mathbf{v}) = \frac{A}{A + K \cdot B}$, 
where $A$ and $B$ represent the overall similarities to ID and OOD prototypes, respectively, and $K$ controls the relative weight of the OOD term.  
When $K$ increases, $A + K \cdot B$ becomes larger, leading to a smaller score. For ID samples ($A \gg B$), moderate increases in $K$ have little effect, but overly large values break the $A \gg B$ assumption and reduce ID scores.  
For OOD samples ($B > A$), increasing $K$ further suppresses the score, but as $K$ grows, the reduction becomes marginal since the score approaches zero. In practice, the optimal range for $K$ is between 5 and 10, while values greater than 10 cause performance degradation due to excessive suppression of ID scores.  

Finally, we study the effect of the cold-start iteration $T_{\text{cold}}$, ranging from $[1,60]$. The performance of our score $S_{\text{proto}}$ quickly surpasses that of the base detector, enabling DynProto to achieve strong performance after only a few cold-start rounds. Overall, DynProto remains stable across a wide range of hyperparameters and consistently outperforms the baseline.

\noindent\textbf{Computation Cost.}
DynProto introduces a modest computational overhead during the caching and clustering stages, which we quantify through empirical measurements. With CLIP-B/16 as the backbone, standard inference runs at 453.3 FPS, whereas our method reaches 428.7 FPS, accompanied by an additional memory cost of 22 MB. With ResNet-50, it decreases from 841.4 to 811.9 FPS with 60 MB memory cost.
\section{Further Discussion}
\begin{figure}[h]
    \centering
    \includegraphics[width=0.5\textwidth]{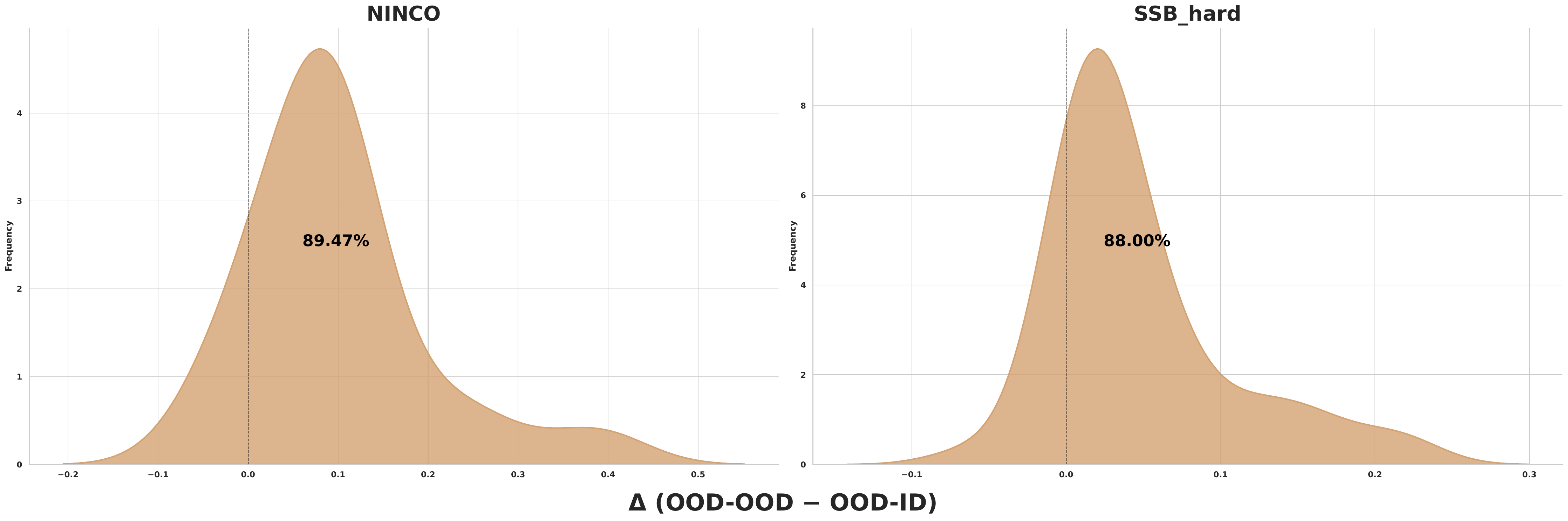}
    \caption{Distribution of per-class similarity difference between OOD-OOD and OOD-ID pairs on CLIP-B/16.   }
    \label{fig:statistics}
\end{figure}

\noindent\textbf{Validation of the Hypothesis.} 
Figure~\ref{fig:statistics} shows the per-class distribution of the difference between the average OOD–OOD and OOD–ID similarities, denoted as $\Delta$, where OOD-OOD denotes the average visual similarity between low-score (detectable) and high-score (undetectable) OOD samples predicted as the same class,  and OOD-ID represents the average similarity between low-score OOD samples and the ID samples of that class. For most classes, OOD samples tend to be more similar, which strongly supports our observation.

\begin{figure}[!h]
    \centering
    \includegraphics[width=0.5\textwidth]{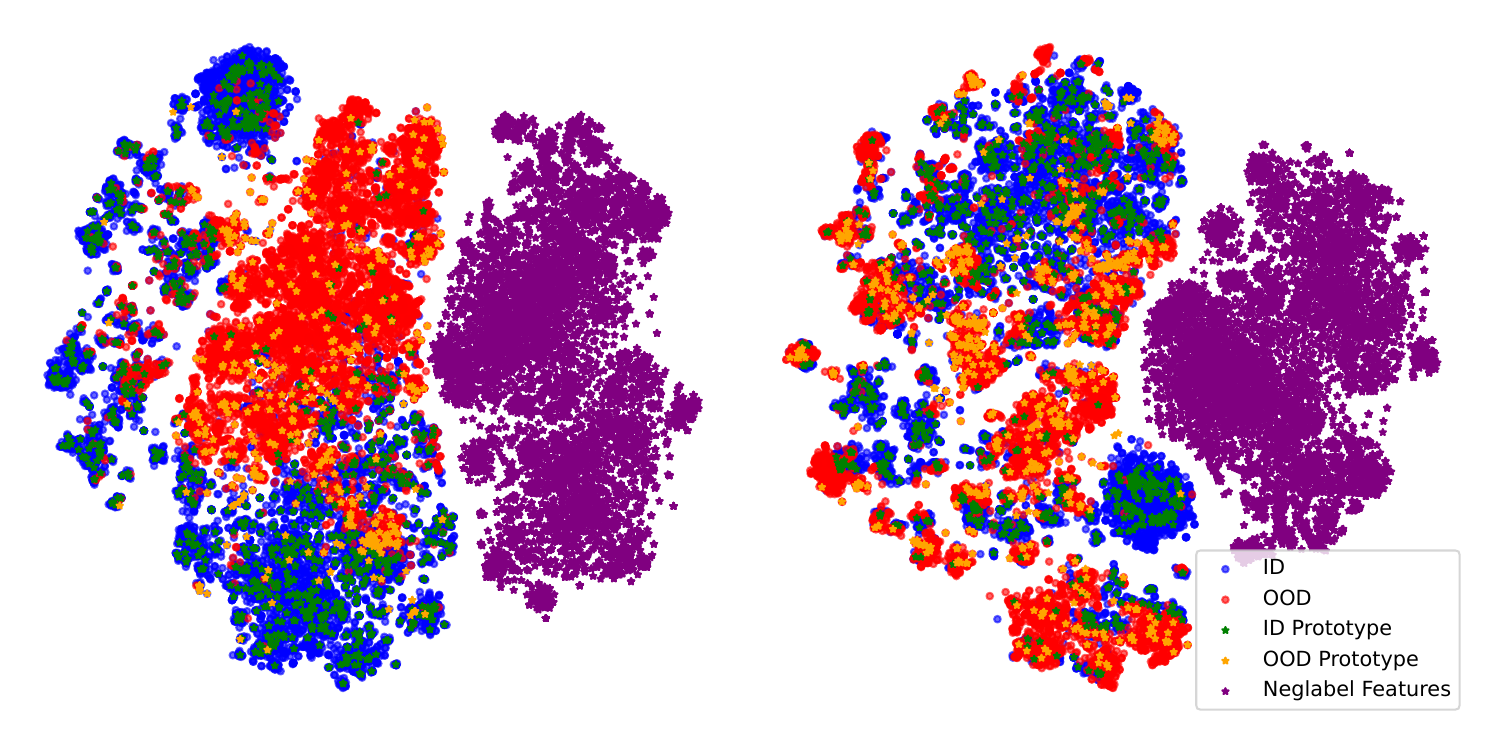}
    \caption{The t-SNE visualization of the constructed OOD prototypes on CLIP-B/16. The ID dataset is ImageNet, while the OOD datasets are Places (left) and SSB-hard (right), respectively.}
    \label{fig:ood_prototypes_visualization}
    \vspace{-1.5em}
\end{figure}
\noindent\textbf{Visualization of the Constructed OOD Prototypes.} As shown in Figure~\ref{fig:ood_prototypes_visualization}, the OOD prototypes constructed by DynProto are well-aligned with the OOD samples, indicating that they effectively capture the underlying visual structure of OOD samples. In contrast, the NegLabel~\cite{neglabel} textual features are misaligned with the visual feature of OOD samples and fail to accurately capture OOD semantics.

\section{Conclusion}
We introduced DynProto, a test-time OOD detection framework that dynamically constructs OOD prototypes during inference, without requiring additional training or predefined OOD labels. DynProto exploits the tendency of OOD samples to cluster in the feature space. It uses easily detectable OOD instances as anchors and identifies harder cases through class-wise caching and clustering-based prototype aggregation. Extensive experiments across a range of OOD benchmarks demonstrate the effectiveness and robustness of our approach. We believe that DynProto can be extended to broader vision tasks, which will be explored in future work.

\section*{Acknowledgments}
This work is supported in part by the National Natural Science Foundation of China (grant No. 62071502), the Major Key Project of PCL (grant No. PCL2023A09),
and Guangdong Excellent Youth Team Program (grant No.2023B1515040025).

{
    \bibliographystyle{IEEEtran}
    \bibliography{main} 
}

\begin{IEEEbiography}[{\includegraphics[width=1in,height=1.25in,clip,keepaspectratio]{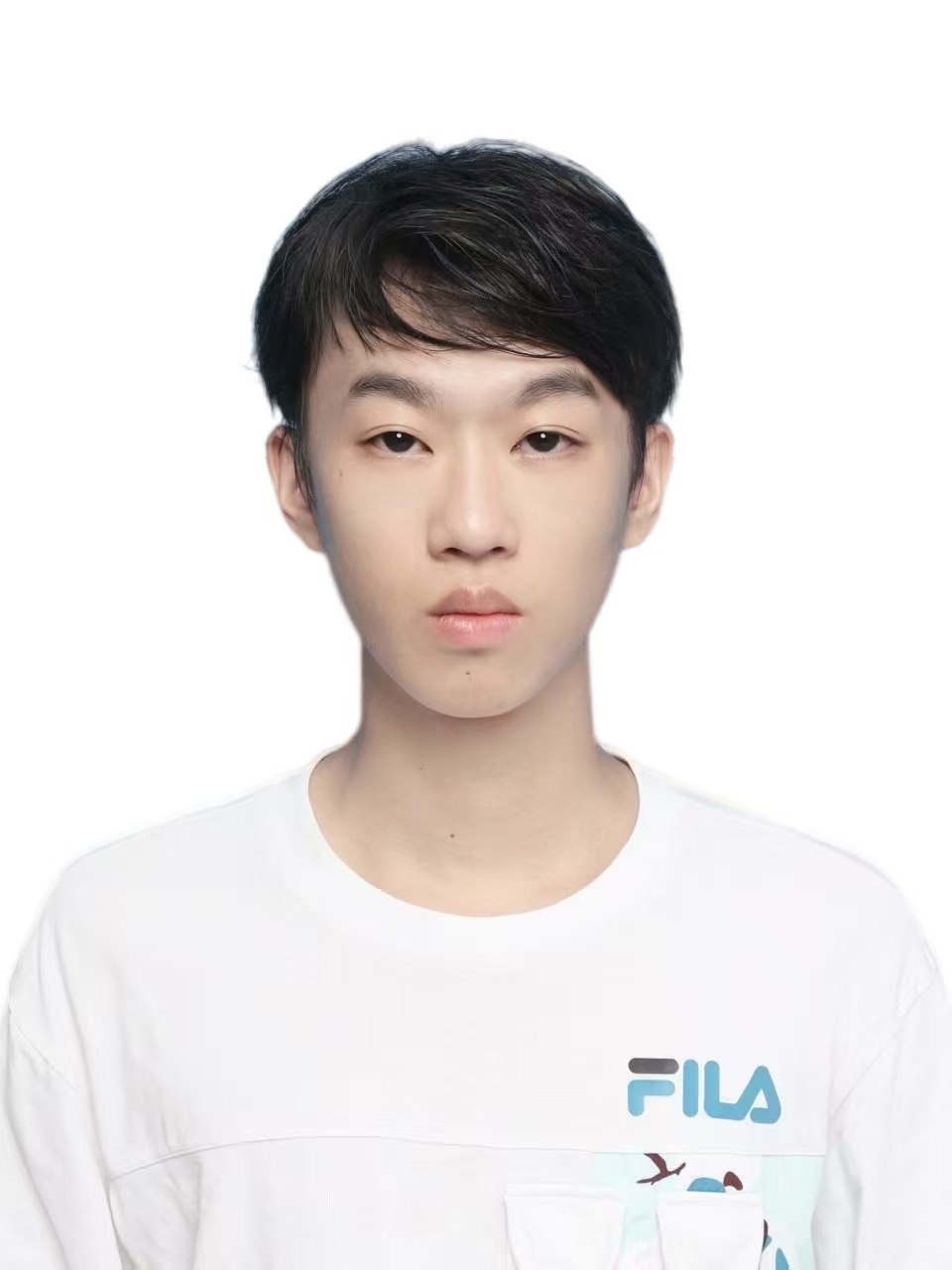}}]{Yanqi Wu} 
received the bachelor’s degree in Network Engineering from South China University of Technology in 2024. He is currently pursuing the master’s degree with the School of Computer Science and Engineering at Sun Yat-sen University. His research interests include computer vision and out-of-distribution detection.
During his studies, he has published one paper as the first author at AAAI, and one papers as a co-author at ICCV.
\end{IEEEbiography}

\vspace{-3em}

\begin{IEEEbiography}[{\includegraphics[width=1in,height=1.25in,clip,keepaspectratio]{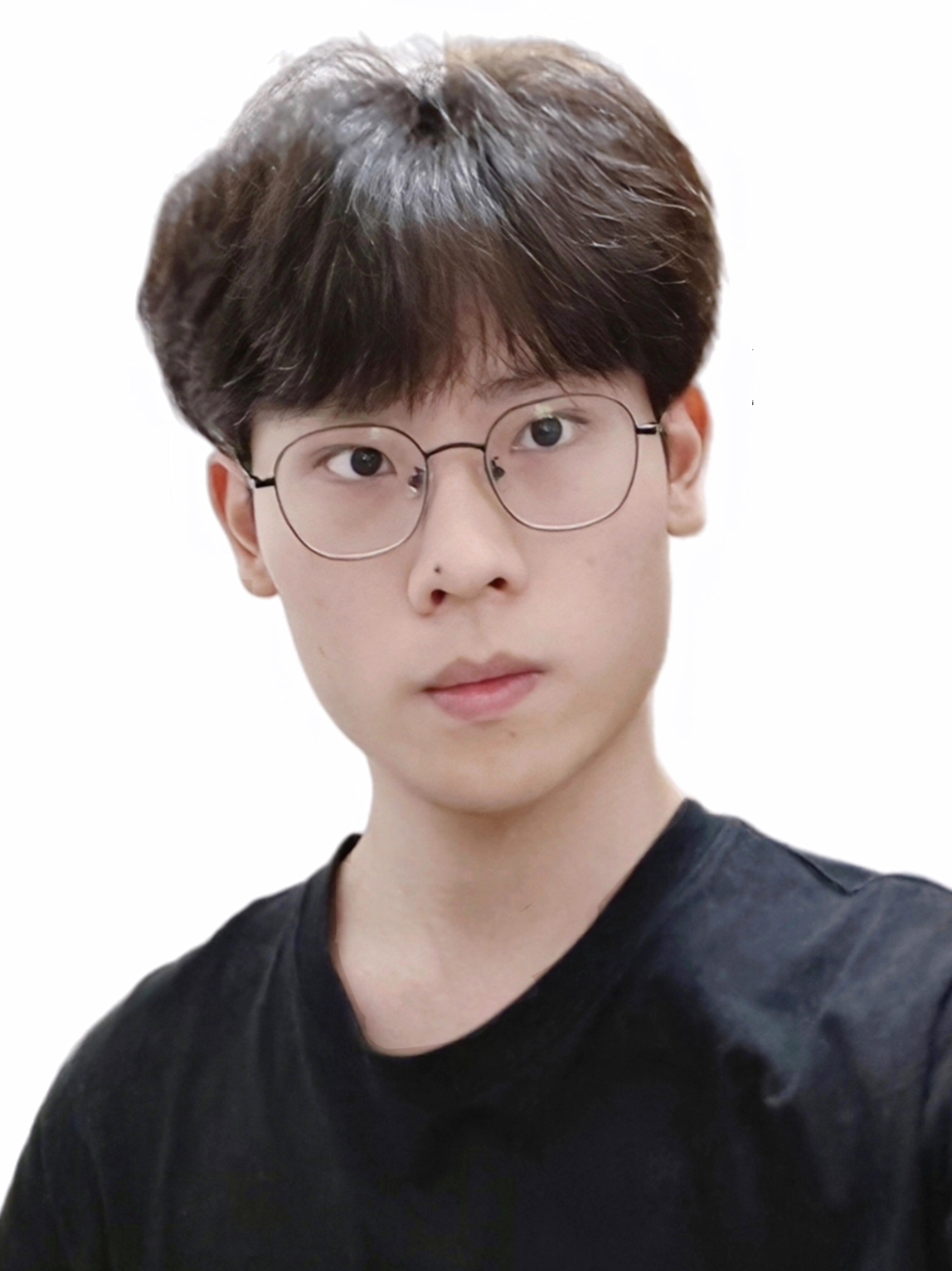}}]{Xinhua Lu} 
received the bachelor’s degree in Network Engineering from South China Normal University in 2024. He is currently pursuing the master’s degree with the School of Computer Science and Engineering at Sun Yat-sen University. His research interests include computer vision and out-of-distribution detection.
During his studies, he has published one paper as the first author at ICCV and coauthored papers at MICCAI, CVPR, and AAAI.
\end{IEEEbiography}
\vspace{-3em}

\begin{IEEEbiography}[{\includegraphics[width=1in,height=1.25in,clip,keepaspectratio]{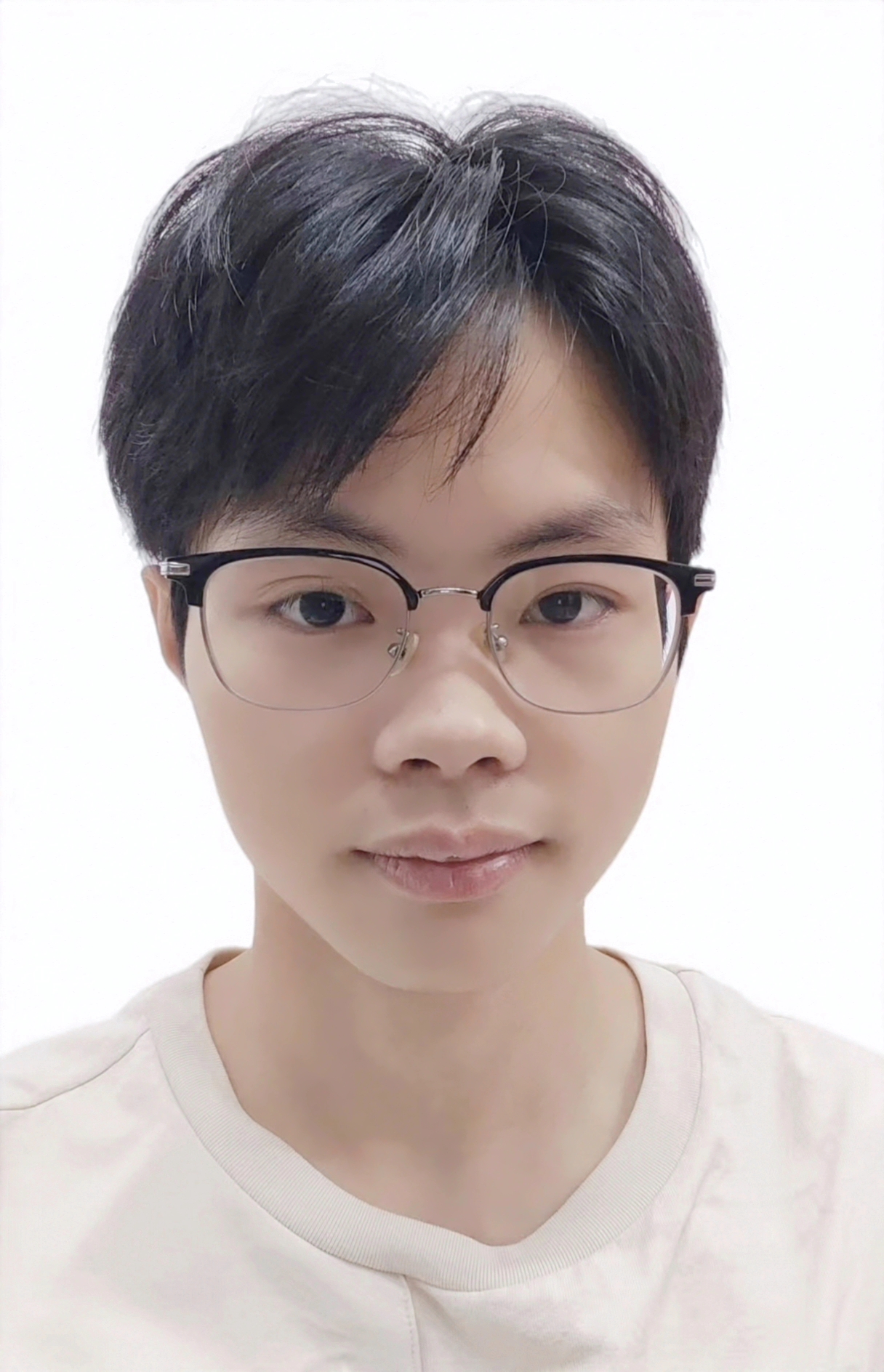}}]{Runhe Lai} 
received the bachelor’s degree in Information and Computing Science from South China University of Technology in 2024. He is currently pursuing the master’s degree with the School of Computer Science and Engineering at Sun Yat-sen University. His research interests include computer vision and out-of-distribution detection.
During his studies, he has published one paper as the first author at MICCAI, and coauthored papers at ICCV, CVPR, and AAAI.
\end{IEEEbiography}
\vspace{-3em}

\begin{IEEEbiography}[{\includegraphics[width=1in,height=1.25in,clip,keepaspectratio]{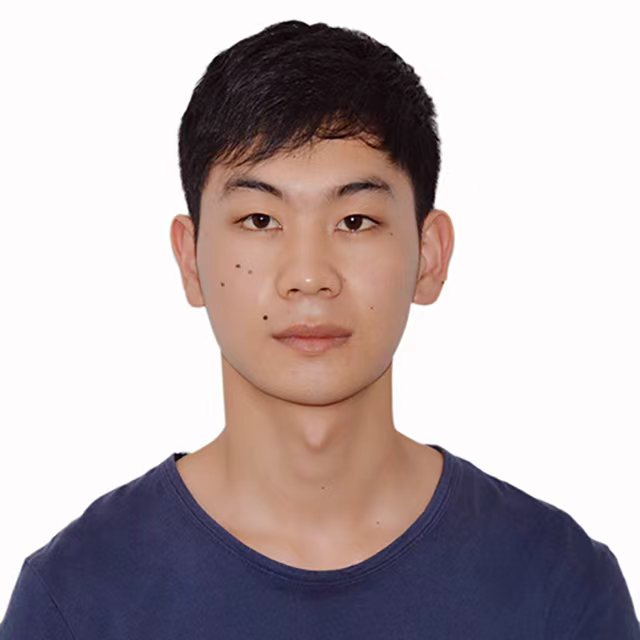}}]{Qichao Chen} is currently pursuing the Ph.D. degree in Computer Science with the University of Nottingham. His research interests include out-of-distribution detection, computer vision, vision-language models, and machine learning. His current work focuses on developing robust and reliable visual recognition systems, with particular emphasis on OOD detection and open-world visual understanding. During his studies, he has coauthored papers at AAAI.
\end{IEEEbiography}
\vspace{-3em}

\begin{IEEEbiography}[{\includegraphics[width=1in,height=1.25in,clip,keepaspectratio]{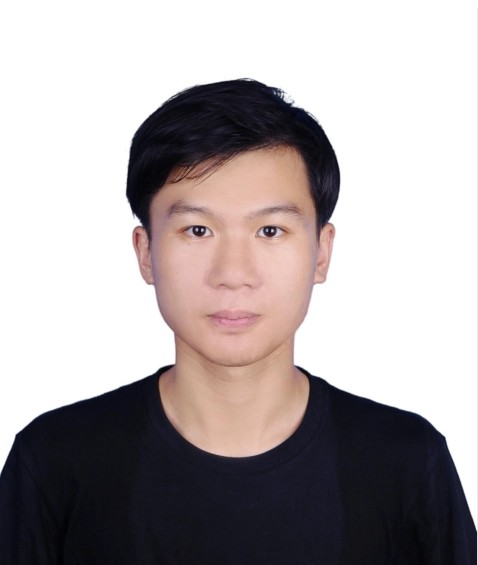}}]{Jia-Xin Zhuang} started his Ph.D. at the Department of Computer Science and Engineering at the Hong Kong University of Science and Technology from 2022, advised by Prof. Hao Chen. He received his M.Eng. as well as B.Eng. degrees at the Department of Computer Science and Engineering of Sun Yat-sen University, advised by Ruixuan Wang, Jianguo Zhang, and Wei-Shi Zheng. His research interests lie in computer vision and medical image analysis, especially self-supervision and foundation models.
\end{IEEEbiography}
\vspace{-3em}

\begin{IEEEbiography}[{\includegraphics[width=1in,height=1.25in,clip,keepaspectratio]{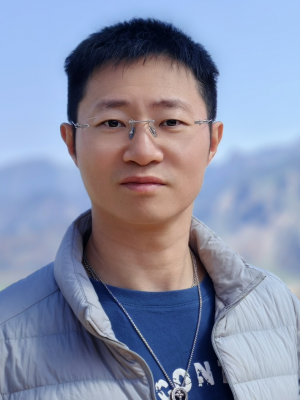}}]{Wei-Shi Zheng} 
is currently a full professor with Sun Yat-sen University. His research interests include person/object association and activity understanding, and the related weakly supervised/unsupervised and continual learning machine learning algorithms. He has now published more than 200 papers, including more than 150 publications in main journals (TPAMI, IJCV, TIP) and top conferences (ICCV, CVPR, SIGGRAPH, ECCV, NeurIPS). He has ever served as area chairs of ICCV, CVPR, ECCV, BMVC, NeurIPS, etc. He serves as an Associate Editor and is on the Editorial Board of IEEE-TPAMI, Artificial Intelligence Journal, Pattern Recognition. He has participated in the Microsoft Research Asia Young Faculty Visiting Programme. He is a Cheung Kong Scholar Distinguished Professor, a recipient of the Excellent Young Scientists Fund of the National Natural Science Foundation of China, and a recipient of the Royal Society-Newton Advanced Fellowship of the U.K.
\end{IEEEbiography}
\vspace{-3em}

\begin{IEEEbiography}[{\includegraphics[width=1in,height=1.25in,clip,keepaspectratio]{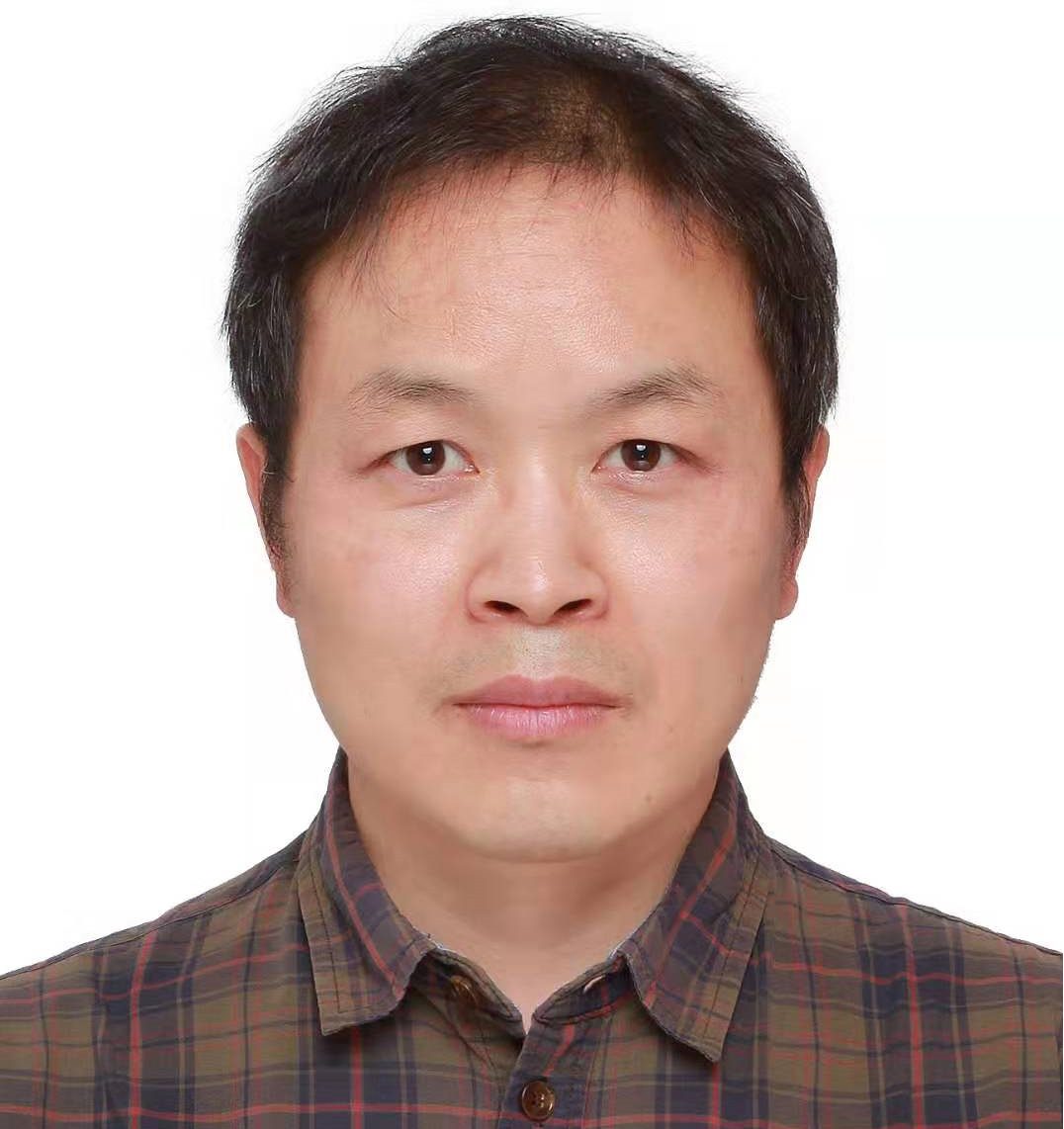}}]{Ruixuan Wang}
is currently a full professor with Sun Yat-sen University. His research interests include computer vision, medical image analysis, and machine learning.
He has published over 80 papers in leading international journals and conferences such as Nature sub-journals, ICCV, CVPR, AAAI, and MICCAI, in the fields of computer vision, pattern recognition, and medical image analysis. He has long served as a reviewer for journals including IEEE TMI, MIA, and TIP, and as a program committee member for conferences such as MICCAI, CVPR, and ICCV. He currently serves on the committees of MICS and the CAA Hybrid Intelligence Committee, and is an executive member of the CCF Technical Committee on Digital Medicine.
\end{IEEEbiography}

\end{document}